\documentclass[3p,times]{elsarticle}

\usepackage[bookmarks=false]{hyperref}

\usepackage{booktabs}
    \hypersetup{colorlinks,
      linkcolor=blue,
      citecolor=blue,
      urlcolor=blue}

\usepackage{amssymb}

\usepackage[figuresright]{rotating}

\begin{document}
\begin{frontmatter}




\title{Latent fingerprint enhancement for accurate minutiae detection}


\author[a]{Abdul Wahab} 
\author[b]{Tariq Mahmood Khan}
\author[c,a]{Shahzaib Iqbal}
\author[d,a]{Bandar AlShammari\corref{cor1} } 
\author[e]{Bandar Alhaqbani}
\author[b,a]{Imran Razzak }

\address[a]{TCC Research and Development Labs, Technology Control Company, Riyadh, Saudi Arabia}
\address[b]{University of New South Wales, Sydney, NSW, Australia}
\address[c]{Abasyn University Islamabad, Islamabad, Pakistan}
\address[d]{Computer and Information Sciences College, Jouf University, Saudi Arabia}
\address[e]{Technology Control Company, Riyadh, Saudi Arabia}

\begin{abstract}

Identification of suspects based on partial and smudged fingerprints, commonly referred to as fingermarks or latent fingerprints, presents a significant challenge in the field of fingerprint recognition. Although fixed-length embeddings have shown effectiveness in recognising rolled and slap fingerprints, the methods for matching latent fingerprints have primarily centred around local minutiae-based embeddings, failing to fully exploit global representations for matching purposes. Consequently, enhancing latent fingerprints becomes critical to ensuring robust identification for forensic investigations. Current approaches often prioritise restoring ridge patterns, overlooking the fine-macroeconomic details crucial for accurate fingerprint recognition. To address this, we propose a novel approach that uses generative adversary networks (GANs) to redefine Latent Fingerprint Enhancement (LFE) through a structured approach to fingerprint generation. By directly optimising the minutiae information during the generation process, the model produces enhanced latent fingerprints that exhibit exceptional fidelity to ground-truth instances. This leads to a significant improvement in identification performance. Our framework integrates minutiae locations and orientation fields, ensuring the preservation of both local and structural fingerprint features. Extensive evaluations conducted on two publicly available datasets demonstrate our method's dominance over existing state-of-the-art techniques, highlighting its potential to significantly enhance latent fingerprint recognition accuracy in forensic applications.

\end{abstract}

\cortext[cor1]{Corresponding author. }
\end{frontmatter}

\email{bmalshammari@tcc-ict.com}



\section{Introduction}
\label{sec:main}
Fingerprints, first used in criminal investigations in 1891, have advanced to become the principal means of quickly and accurately identifying perpetrators \cite{caplan1990fingerprints}. Their widespread use in legal proceedings worldwide has consolidated their status as key technological proof in criminal prosecutions. In many cases, latent prints, which are unintentionally left on surfaces such as metal, plastic, and glass, may evade detection without the help of specialised tools. To reveal these prints, specialised techniques are used to lift and analyse them, improving the visibility of the friction ridge patterns they contain \cite{khan2010fingerprint, khan2013fingerprint, khan2014fingerprint, khan2016spatial}. Subsequently, the enhanced images can be verified by comparing them with a repository of previously recorded ten-print fingerprints (that is, slapped or rolled) in law enforcement data, facilitating the identification of the individual responsible for the prints \cite{maltoni2009handbook, razzak2010multimodal, razzak2011multimodal}. Latent print recognition has been integral to forensic analysis and the pursuit of justice in criminal investigations for decades, providing credible information for suspect identification. However, the reliance on latent to rolled fingerprint matching trails behind rolled-to-rolled matching, culminating in instances such as inappropriate bound due to inaccurate comparisons by automated fingerprint identification systems (AFIS) and failures to adhere to the ACE-V (Analysis, Comparison, Evaluation, and Verification) technique \cite{ashbaugh1999quantitative} as exemplified in cases like \cite{oig2006review}. The challenges inherent in latent fingerprint recognition include low valley-ridge contrast, occlusion, distortion, diverse backgrounds, and incomplete patterns, making it a multifaceted conundrum in biometrics.

AFIS is primarily optimised for high-quality rolled or plain fingerprint impressions. However, their accuracy demonstrably degrades when presented with latent fingerprints, even after refinements to publicly available datasets. This performance disparity has motivated extensive research into specialised fingerprint recognition techniques optimised for the specific challenges inherent in latent fingerprint analysis. These studies concentrate on various areas such as extracting minutiae \cite{tang2017latent, khan2016stopping, khan2016efficient, khan2017efficient,khan2018coupling,sabir2020reducing,khan2022hardware,khan2022fusion}, latent enhancement \cite{li2018deep, huang2020latent, zhu2023fingergan, joshi2019latent}, foreground segmentation of ridge structures \cite{cao2014segmentation}, and estimation of the orientation field \cite{feng2012orientation, siddiqui2024robust}. Despite advances in individual fingerprint enhancement techniques, research on integrating them into a complete latent-to-rolled fingerprint recognition system remains limited. This is crucial because poorly integrated components can hinder overall system performance. While some end-to-end recognition systems have been proposed \cite{tang2017fingernet, cao2018automated, cao2019end}, the best reported rank-1 identification rate using the NIST SD27 dataset (258 latent probes vs. 100,000 rolled fingerprints) is only 65\%. This highlights the need for further research on optimising the combined performance of all components within a holistic latent fingerprint recognition system.

Deep learning has emerged as a powerful tool in recent years, revolutionising a wide range of applications by leveraging its ability to learn complex patterns from vast amounts of data \cite{tabassum2020cded,khan2020exploiting,khan2020semantically, khan2021residual, khan2022t, iqbal2022g,arsalan2022prompt,khan2023retinal}. Deep learning advances have produced effective fixed-length embeddings for fingerprint recognition. However, applying these global representations to latent prints poses challenges due to inherent domain gaps and the limited availability of extensive latent fingerprint datasets required for training. To address this challenge, our study proposes an integrated end-to-end pipeline for latent fingerprint recognition. This pipeline integrates a learnt global fingerprint representation, which highlights ridge patterns, with local representations such as minutiae. Through this integration, our objective is to improve the accuracy and search efficiency in latent-to-rolled fingerprint recognition tasks. By integrating these complementary local and global features, our approach leverages local characteristics to guide global representations towards emphasising distinctive regions within the input fingerprint image pairs, thereby resulting in significantly improved identification accuracy. Unlike existing latent-to-rolled fingerprint matching methods, our proposed representation and matching pipeline demonstrates versatility and effectiveness across diverse fingerprint image types, encompassing plain, rolled, latent, and even contactless acquisitions. 

Moreover, it accommodates various fingerprint sensors, such as optical and capacitive technologies. Basically, we propose the development of a universal fingerprint representation that transcends the limitations imposed by the varying sensor types and the diverse modes of fingerprint capture. This representation serves as a versatile framework capable of accommodating a wide range of fingerprint data, regardless of the specific sensor technology employed or the method used to capture the fingerprints. By adopting this approach, we aim to establish a standardised representation that ensures consistency and interoperability between different fingerprint systems, thus facilitating seamless integration and interoperability in various applications, including authentication, identification, and forensic analysis.

This paper presents the following notable contributions:
\begin{itemize}
    \item We propose a novel approach that leverages Generative Adversarial Networks (GANs) to revolutionize Latent Fingerprint Enhancement (LFE) by employing a structured method for fingerprint generation.
    \item By directly optimising minutiae information during the generation process, our model produces enhanced latent fingerprints with exceptional fidelity to ground-truth instances.
    \item Assessing the model's ability to accurately recover minutiae, essential fingerprint features crucial for identification.
    \item Evaluating the performance of the proposed method in comparison to the current state-of-the-art model.
\end{itemize}

The remainder of this paper is structured as follows. Section \ref{sec:LR} offers a comprehensive review of the literature. Section \ref{sec:method} details the proposed method. Section \ref{sec:rd} presents the experimental results. Finally, Section \ref{con} provides the conclusion of the paper.

\section{Literature Review}
\label{sec:LR}
Fingerprints are widely used for human authentication and identification, playing a vital role in civil and criminal applications \cite{maltoni2009handbook, hawthorne2017fingerprints}. In recent years, significant efforts have been directed towards the enhancement of latent fingerprints \cite{abebe2020latent, malwade2015survey, wallace2004detection, sankaran2014latent, schuch2018survey}. In the initial phases, conventional image processing methods, such as contextual and directional filtering, were employed for this purpose. For example, Cappelli et al. \cite{cappelli2009semi} introduced the idea of adjusting a gabor filter to correspond to the local orientations and frequencies of fingerprints, in order to reduce noise and enhance clarity in ridge structures. Similarly, this study \cite{chikkerur2007fingerprint} suggested employing contextual filtering in the Fourier domain to improve fingerprint quality. However, while these techniques demonstrate proficiency in enhancing low-quality plain or rolled fingerprints, they often encounter challenges when applied to latent fingerprints, primarily due to two key factors: 1) the presence of distorted ridge structures due to structural noise and 2) Blurred ridge structures lead to unreliable orientation and frequency estimation.

The enhancement of the estimation of orientation in latent fingerprints has been a subject of considerable research, leading to the proposal of various global smoothing and modelling techniques \cite{yoon2010latent, yoon2011latent, feng2012orientation}. Yoon et al. \cite{yoon2011latent} proposed a methodology that combines a polynomial model with Gabor filters to achieve a more precise orientation estimation. Building on this work, Feng et al. \cite{feng2012orientation} suggested the use of an orientation patch dictionary in conjunction with Gabor filtering for LFE. Expanding on these ideas, Yang et al. \cite{yang2014localized} introduced an improved approach by substituting the orientation dictionary with localised orientation dictionaries. These dictionaries vary depending on their spatial positions, enabling more robust orientation estimation and consequently leading to improved print enhancement outcomes. However, despite these advances, these methods may encounter challenges in real-world applications where fingerprint ridge frequencies exhibit inconsistencies. A fundamental challenge when using Gabor filters lies in their inherent reliance on a fixed ridge frequency during filter tuning, which may not accurately reflect the variability present in actual fingerprint data. Thus, while these techniques represent significant progress in LFE, their efficacy in practical scenarios may be hindered by such limitations.

In pursuit of further enhancing latent fingerprints, researchers have turned to total variation (TV) image models. These models operate by minimising the total variation of an image, effectively breaking it down into texture and cartoon components, thus harnessing ridge structures. For instance, \cite{zhang2012latent} introduced an adaptive TV model aimed at removing structural interference in latent fingerprints for accurate identification. Subsequently, they introduced an adaptive directional TV model aimed at enhancing latent prints. These methods showcase effectiveness in reducing structural noise within the texture components of latent fingerprints by considering local orientations and scales. However, a significant challenge arises in accurately estimating local parameters for low-quality latent fingerprints, resulting in extracted ridge structures that are often weak. As a result, in later research, TV decomposition was often used as a preprocessing technique to improve latent prints \cite{liu2014latent}.

Consequently, the emergence of deep learning has revolutionised various tasks related to fingerprints, with deep neural networks playing a crucial role in LFE \cite{darlow2017fingerprint, nguyen2018robust}. For example, Cao et al. proposed a two-stage CNN-based approach: first estimating the orientation field and then refining the fingerprint with Gabor filters \cite{cao2015latent}. Svoboda et al. investigated a different approach using a convolutional autoencoder network specifically designed for latent fingerprint reconstruction \cite{svoboda2017generative}, similarly, Li et al. \cite{li2018deep} introduced a deep convolutional network comprising both convolutional and deconvolutional layers to enhance latent fingerprints. Horapong et al. explored the application of sparse autoencoders specifically to enhance the ridge / valley structure \cite{horapong2020progressive}. While \cite{liu2020automatic} introduced a deep-nested UNet architecture specifically designed to address the challenges of latent fingerprints. Despite the promising results presented by these methodologies, the persistent challenge of restoring damaged ridge and valley structures with latent fingerprints continues to be a prevalent issue in numerous approaches.

Recent advances have witnessed the emergence of GANs \cite{naqvi2023glan} as a transformative tool in LFE. GANs have shown significant progress in recovering critical ridge and valley structures that are fundamental for fingerprint identification \cite{huang2020latent, zhu2023fingergan, joshi2019latent, dabouei2018id}. A pioneering contribution in this field is the work of Dabouei et al., who introduced a condition-specific GAN architecture specifically tailored to enhance partial latent fingerprints. This approach not only demonstrably improves ridge clarity in relatively well preserved areas but also effectively excludes severely damaged regions, showcasing the potential of GANs for selective enhancement and mitigating the impact of data imperfections \cite{dabouei2018id}. Building upon this foundation, subsequent research by Zhu et al. and Joshi et al. explored further refinements to GAN-based algorithms, achieving even more remarkable results in enhancing the inherent ridge / valley structure of the fingerprint \cite{zhu2023fingergan, joshi2019latent}. Furthermore, Huang et al. introduced a PatchGAN methodology for LFE, thus contributing another similar approach to the domain explored in \cite{huang2020latent}.

\section{Proposed Methodology}
\label{sec:method}

To provide a clear understanding of our proposed method, this section outlines the step-by-step workflow for latent fingerprint enhancement. We used a visual representation in Fig. ~\ref{flow} to illustrate each stage of the process. This will allow readers to easily follow the logic behind our approach and to get a comprehensive picture of how latent fingerprints are enhanced. Before diving into the network architecture and objective function, it is crucial to grasp some fundamental aspects of GANs. Initially, the discriminator undergoes training with real data, ensuring its proficiency in discerning between real and fake input. The output of the discriminator serves as input for the generator, which iteratively refines the data to effectively deceive the discriminator. The objective is to improve the quality of noisy latent fingerprints, preserving their distinctive features while removing noise. Based on unsupervised learning principles, the machine automatically identifies patterns within input images, striving to retain maximal fidelity. This adaptive learning approach empowers the generator network to generate outputs that closely resemble the original images.

\subsection{Pre-processing}

\begin{figure*}[t]\vspace*{4pt}
    \centering\includegraphics{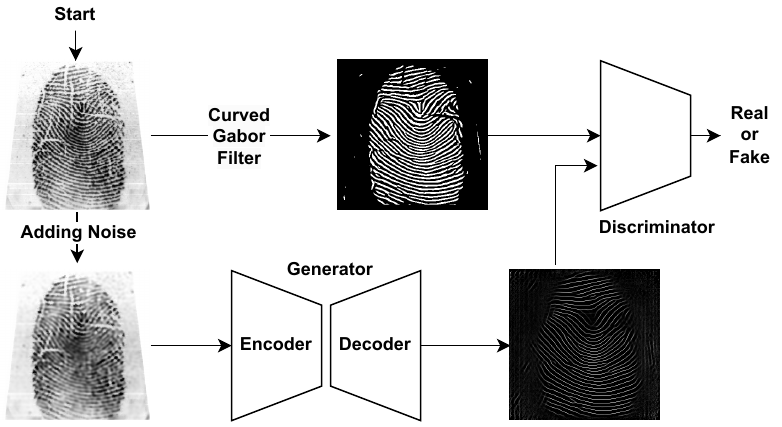}
    \caption{Framework for latent fingerprint enhancement and reconstruction}
    \label{flow}
\end{figure*}

In the realm of latent fingerprints, the integration of deep learning presents significant challenges due to deficiencies in existing public datasets. These datasets often lack sufficient correlation between latent fingerprints and their genuine matches or suffer from inadequacies in quantity. To address this, we propose a novel approach to generate training data, as depicted in Figure \ref{flow}. Initially, we used a curved Gabor filter \cite{gottschlich2011curved} to enhance the ridges and valleys of the latent input fingerprint in various orientations, which is then passed to the discriminator as the truth of the ground. To simulate realistic noise and enhance the diversity of training data, we incorporate complex noise patterns proposed in prior works \cite{zhu2023fingergan, shreya2023gan}; rather than simple line or character noise. This enhancement strategy enables the generation of extensive training data that closely resemble genuine latent fingerprint scenarios, thus facilitating more effective learning by the proposed generator network under challenging conditions. The resulting decomposition is then fed into the generator network. During the process, noise is introduced into the input fingerprints to generate synthesised latent fingerprints that are used as ground truth data by the discriminator.

\subsection{Generator Architecture}

The generator network takes a latent fingerprint as input, aims to enhance it, and produces the output as an enhanced latent fingerprint (Fig. \ref{flow}). In the proposed implementation, we have employed four encoder-decoder blocks. Let $l^{n\times n}$ be the $n\times n$ convolution operation $f^{n\times n}$ followed by batch normalisation ($\beta _{n}$) and ReLU ($\Re$) operations for any given input ($\texttt{In}$) as defined by (Eq. \ref{Eq1}).

\begin{equation}
l^{n\times n} = \Re \left ( f^{n\times n}\left ( \texttt{In} \right ) \right )
    \label{Eq1}
\end{equation}

The initial skip connection ($s_{o}$) is computed by applying the $l^{3\times 3}$ operation to the input of the network ($X_{in}$) as shown in (Eq. \ref{Eq2}).

\begin{equation}
    s_{o}=l^{3\times 3}(X_{in})
    \label{Eq2}
\end{equation}

Similarly, the output of the initial encoder block denoted by ($E_{o}$) is computed as (Eq. \ref{Eq3}).

\begin{equation}
E_{o}=m_{p}\left ( l^{3\times 3}\left ( l^{3\times 3}\left (s_{o}  \right ) \right ) \right )
\label{Eq3}
\end{equation}

where ($m_{p}$) is the maxpooling operation. The output of the encoder block $k^{th}$ ($E_{k}$) is computed by (Eq. \ref{Eq4}).

\begin{equation}
    E_{k}=m_{p}\left [ \Re\left \{ \beta _{n}\left ( f^{3\times 3}\left ( \beta _{n}\left ( f^{3\times 3}\left ( s_{k} \right ) \right ) \right ) \right ) + f^{3\times 3}\left ( l^{3\times 3}\left ( l^{3\times 3} \left ( E_{k-1} \right )\right ) \right )\right \} \right ]
    \label{Eq4}
\end{equation}

where ($s_{k}$) is the $k^{th}$ skip connection and is computed as given in (Eq. \ref{Eq5}).
\begin{equation}
    s_{k}=l^{3\times 3} (E_{k-1})
    \label{Eq5}
\end{equation}

Once the information is extracted by the encoder block, it is given to the decoder stage to reconstruct the spatial feature maps.

\begin{equation}
    \Im _{k} = (s_{k}) + l^{3\times 3}(u_{p}(D_{k-1}))
    \label{Eq7}
\end{equation}
where $u_{p}$ is the upsampling operation that increases the size of the feature maps in terms of their spatial characteristics. The output of the $k^{th}$ decoder block is computed using (Eq. \ref{Eq8}).

\begin{equation}
    D_{k}=\Re \left [ f^{3\times 3}\left ( l^{3\times 3}\left ( l^{3\times 3}\left ( u_{p}\left ( D_{k-1} \right ) \right ) \right ) \right )+\beta _{n}\left ( f^{3\times 3}\left ( \beta _{n}\left ( f^{3\times 3}\left ( \Im _{k} \right ) \right ) \right ) \right ) \right ]
    \label{Eq8}
\end{equation}

The output of the model ($X_{out}$) is computed by applying the $l^{3\times 3}$ operation followed by the ($f^{1\times 1}$) convolution and the sigmoid ($\sigma$) operation as shown in (Eq. \ref{Eq9}).

\begin{equation}
    X_{out}= \sigma(f^{1\times 1}(l^{3\times}(\Im _{k})))
    \label{Eq9}
\end{equation}

The generated enhanced latent fingerprint of size $256\times 256$ is obtained in the model output.

\subsection{Discriminator Architecture}

The proposed discriminator architecture is based on a CNN design that features seven composite blocks. The discriminator network uses a sequence of convolutional layers, each of which is followed by batch normalisation for enhanced training stability. Leaky ReLU activation functions are used throughout the network, except for the final layer, which uses a sigmoid function to generate a binary classification score. This architectural design aligns with the parameter choices made for the generator network, particularly the use of small kernels. To prepare input for the discriminator, the generator network generates enhanced latent fingerprints. The fusion of a latent noised ground truth fingerprint with the output of the generator in two distinct two-channel inputs. The primary goal of the discriminator is to distinguish between these inputs, aiming to produce improved representations that exhibit high similarity to the ground-truth fingerprints. This iterative process allows the generator network to cultivate a profound semantic understanding, enabling it to master the restoration of distorted ridge structures in latent fingerprints and perform denoising effectively.

\section{Experimental Results}
\label{sec:rd}
This section assesses the performance of our proposed method. We begin by providing a description of the datasets used for experiments and evaluation, followed by a detailed description of the implementation specifics.

\subsection{Datasets}
The experiments are carried out using the FVC2002 \cite{maio2002fvc2002} database sourced from biometric systems. Within the FVC2002 database, the DB\_1 dataset employs the TouchviewII optical sensor by Idenix, while the DB\_2 dataset utilises the FX2000 optical sensor by Biometrika. The dataset comprises a total of 100 fingerprint images, with 8 samples of each image.

We also used the CASIA latent fingerprint dataset \cite{CASIA_dataset}, collected by the Institute of Automation at the Chinese Academy of Sciences (CASIA), which offers a comprehensive repository of real-world fingerprint images. Using data from crime scenes, forensic databases, and various practical scenarios, this dataset encompasses a wide array of conditions and challenges commonly encountered in fingerprint identification tasks. Each image is typically accompanied by rolled fingerprints or pertinent metadata. With high-resolution capabilities, it facilitates detailed analysis, capturing partial prints, smudges, distortions, and varying image qualities in different scenarios. In addition, CASIA-FingerprintV5 \cite{CASIA_dataset}, provided by the Ideal Biometrics Test, constitutes another fingerprint database within the CASIA collection. This dataset comprises 20,000 fingerprint images sourced from 500 subjects. Each subject contributed 40 fingerprint images, covering eight fingers (thumb, index finger, middle finger, and ring finger for both hands), with five images captured per finger. The acquisition process involved applying various levels of pressure during fingerprint scanning.

\subsection{Implementation Details}
Training a GAN for fingerprint enhancement requires a large amount of paired training data. Ideally, these data would comprise latent fingerprints, considered the ground truth, alongside their corresponding high-quality enhanced counterparts. However, there are no publicly available databases that contain such paired low-quality latent and high-quality enhanced fingerprint images suitable for training purposes. To avoid data scarcity, we propose the generation of synthetic latent fingerprints that emulate real-world acquisition artefacts, including overlapping patterns and intricate backgrounds. This approach involves the creation of synthetic fingerprints, followed by the subsequent enhancement of ground-truth images obtained from curated datasets such as FVC2002 and CASIA-FingerprintV5.

The architectural details of the proposed model, which is specifically designed to enhance latent fingerprints to enable accurate minutiae detection, are illustrated in Fig. \ref{flow}. For the implementation of the proposed model, we use the PyTorch framework, employing the Adam SGD optimiser \cite{kingma2014adam} with a learning rate set to 0.001. The training data was resized to $192 \times 192$, serving as input to train the proposed model. During inference, a sliding window $192 \times 192$ with a batch size of 8 was used to enhance the latent fingerprint. The experiments were carried out on a system equipped with an Intel(R) Xeon (R)320 E5-2630 v4 CPU operating at 2.2–3.1 GHz, 64 GB of RAM, and a Nvidia Titan X (Pascal) GPU with 12 GB of memory.

\subsection{Results and Discussions}
\begin{figure}[t]
    \centering
    \includegraphics[width=0.9\textwidth, height=0.4\textheight]{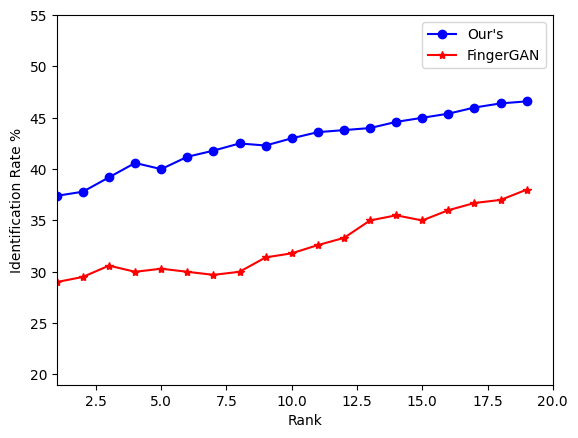}
    \caption{CMC curve presenting the identification rates of enhanced latent fingerprints of the proposed and FingerGAN models.}
    \label{fig:cmc}
\end{figure}

The aim of enhancing latent fingerprints is to improve the clarity of pertinent details while reducing noise artefacts in the image, thereby enhancing the accuracy of recognition. As a result, evaluating the effectiveness of our model can be achieved by evaluating fingerprint identification accuracy.
For fingerprint identification and matching, we use the commercial software VariFinger SDK12.1, available at \url{https://www.neurotechnology.com/}. Using VariFinger allows for a direct comparison with previous studies that also used this software, allowing a fair evaluation of our proposed model against existing methods within the same framework. To assess the performance of latent fingerprint identification, we use the cumulative match characteristic (CMC) curve. The comparison of the CMC curves generated by our model and FingerGAN \cite{zhu2023fingergan} using enhanced latent fingerprints is illustrated in Figure \ref{fig:cmc}. Our method notably achieved the highest rank with an accuracy of 48\%, surpassing FingerGAN's 35\%.

\begin{figure}[t]\vspace*{4pt}
    \centering\includegraphics[width=0.9\textwidth]{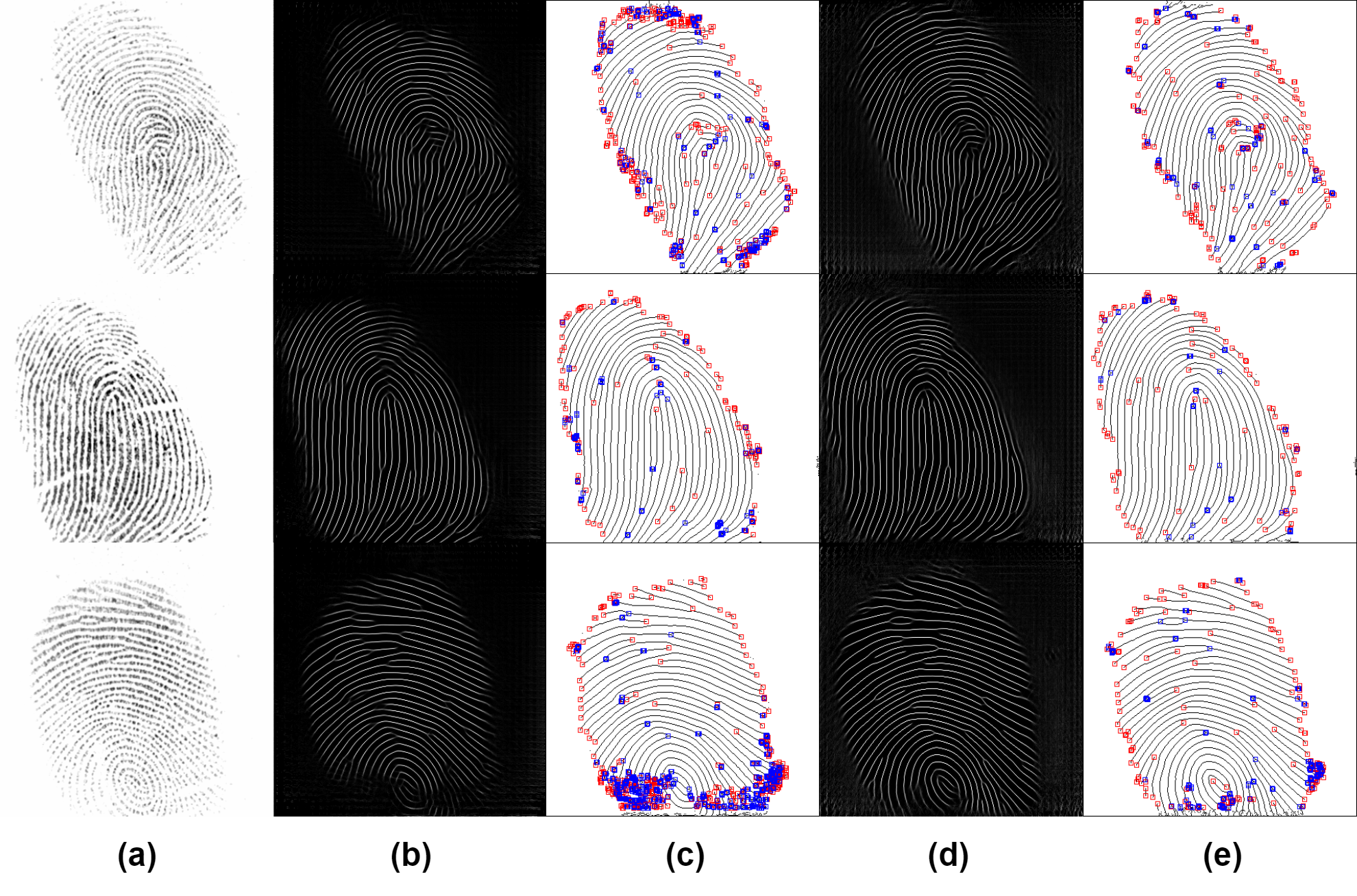}
    \caption{Comparison of the proposed method and FingerGAN on FVC2002-DB1A \cite{maio2002fvc2002} dataset: (a) Ground truth latent fingerprint; (b) FingerGAN enhanced fingerprints; (c) Minutiae extracted from FingerGAN output; (d) Our enhanced fingerprints; (e) Minutiae from our enhanced fingerprint. Ridge endings are in red, and bifurcations are in blue.}
    \label{db100}
\end{figure}

\begin{figure}[t]\vspace*{4pt}
    \centering\includegraphics[width=0.9\textwidth]{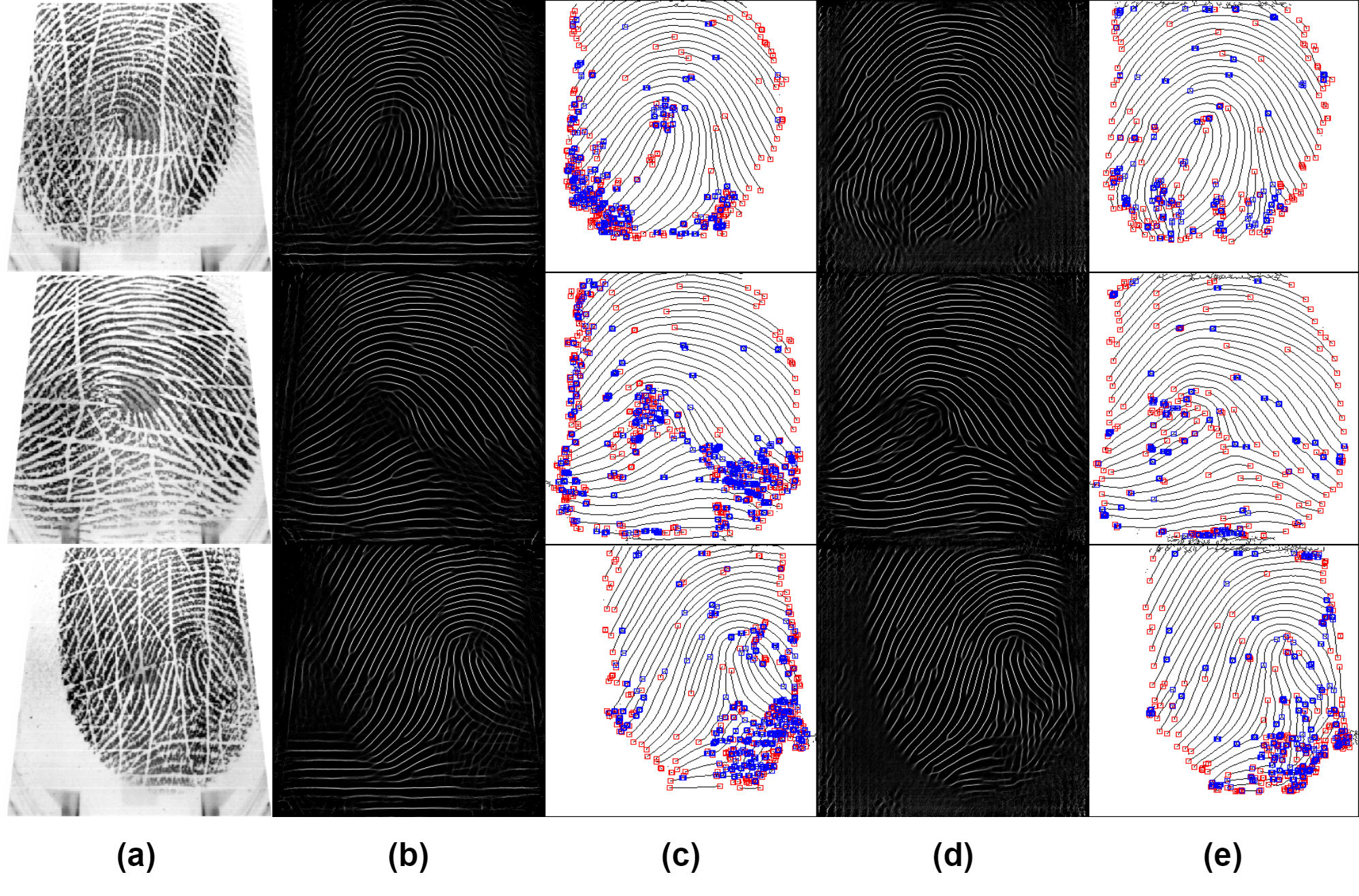}
    \caption{Comparison between our approach and FingerGAN on CASIA \cite{CASIA_dataset}: (a) Ground truth latent fingerprint; (b) FingerGAN enhanced fingerprint; (c) Minutiae extracted from (b); (d) Enhanced fingerprint from our method; (e) Minutiae extracted from (d). Ridge endings are in red, and bifurcations are in blue.}
    \label{casia}
\end{figure}

Furthermore, we explored the accuracy of minutiae recovery through a comparative analysis with a leading method, namely FingerGAN, to gauge the effectiveness of our approach in latent fingerprint enhancement. Initially, genuine minutiae were extracted from ground truths. Following enhancement, VeriFinger 12.1 was used to extract minutiae of enhanced fingerprints for further evaluation. The extracted minutiae were then compared with genuine minutiae to identify recovered minutiae that aligned in terms of precise location, minutia type, and orientation. Any minutiae failing to meet these criteria are classified as fake and attributed to the model. Table \ref{min_tab} presents the quantitative results of the accuracy of the recovery of minutiae. It is evident from the table that our model outperforms FingerGAN by recovering more genuine minutiae while generating fewer fake ones. For a detailed examination of test data indexes concerning enhanced fingerprints. These results underscore the better performance of the proposed model in minutiae recovery accuracy, as visually demonstrated in FigsFig.. \ref{db100} and \ref{casia}, which prominently showcase direct optimisation of minutiae informationFig.

\begin{table}[h]
\caption{Assessment of minutiae recovery accuracy in comparison to FingerGAN, considering the total count of genuine minutiae recovered and fake minutiae introduced for latent fingerprints in the CASIA database.}
\begin{tabular*}{\hsize}{@{\extracolsep{\fill}}lll@{}}
\toprule
Methods & FingerGAN & Our's\\
\toprule
Genuine minutiae recovered &   1431 &  \textbf{1982}\\

Fake minutiae introduced  & 11039 &  \textbf{8361}\\
\bottomrule
\end{tabular*}
\label{min_tab}
\end{table}

A key feature of our model is its ability to generate delicate minutiae points within latent enhanced fingerprints, accurately reflecting those found in the ground truths. In contrast, the current state-of-the-art method struggles to capture these minutiae and often misclassifies them as bifurcations. Furthermore, our model consistently maintains the ridge structures.


\section{Conclusion}
\label{con}
This paper highlights the limitations of current latent fingerprint identification systems, particularly their sensitivity to noise and poor image quality. These limitations demonstrably lead to low identification accuracy. However, a promising solution emerges with the application of a GAN-based approach. The proposed method achieves a statistically significant increase in identification accuracy with an accurate recovery of minutiae and ridge patterns. Enhanced images yield a remarkable accuracy of 43. 6\%, compared to only 21.7\% for raw images. This underscores the potential of deep learning to revolutionise forensic science and facilitate more effective criminal investigations. Further refinement of these methods has immense potential for the development of even more accurate and reliable latent fingerprint identification systems. This, in turn, would significantly enhance the robustness of forensic analysis. 

Future research directions could focus on the selection and comparison of diverse datasets that closely resemble real-world latent fingerprint data. Such an approach would enable for a more rigorous investigation of the generalisability of the reported results.









\bibliographystyle{elsarticle-num}
\bibliography{ref}

\clearpage

\normalMode

\end{document}